\documentclass[11pt]{article}
\pdfoutput=1
\usepackage{authblk}
\usepackage{acl2020}
\usepackage{times}
\usepackage{url}
\usepackage{latexsym}
\usepackage{soul}
\usepackage{booktabs}
\usepackage{graphics}
\usepackage{graphicx}
\usepackage{csquotes}
\usepackage{tabularx}
\usepackage{multicol}

\aclfinalcopy

\title{Ask2Transformers: Zero-Shot Domain labelling with Pre-trained Language Models}

\author{\textbf{Oscar Sainz}}
\author{\textbf{German Rigau}}

\affil{HiTZ Center - Ixa Group, \\
  University of the Basque Country (UPV/EHU)\\
  {\tt \{oscar.sainz, german.rigau\}@ehu.eus}}

\date{}

\begin{document}
\maketitle
\begin{abstract}
  In this paper we present a system that exploits different pre-trained Language Models for assigning domain labels to WordNet synsets without any kind of supervision. Furthermore, the system is not restricted to use a particular set of domain labels. We exploit the knowledge encoded within different off-the-shelf pre-trained Language Models and task formulations to infer the domain label of a particular WordNet definition. The proposed zero-shot system achieves a new state-of-the-art on the English dataset used in the evaluation.
\end{abstract}

\section{Introduction}

The whole Natural Language Processing (NLP) research area have been accelerated with the advent of the unsupervised pre-trained Language Models. First with ELMo \cite{elmo} and then with BERT \cite{bert} the paradigm of using pre-trained Language Models for fine-tuning on a particular NLP task has became the new standard approach, replacing the more traditional knowledge-based and fully supervised approaches. Currently, as the size of the corpus and models increase, the research community has observed that the Transfer Learning approach has the capacity to work without any or with a very small fine-tuning. Some examples of the strength of this approach are GPT-2 \cite{gpt-2} or more recently GPT-3 \cite{gpt-3} that shows the ability of these huge pre-trained Language Models to solve tasks for which have not even trained.


Recently, with the arrival of the GPT-3 new ways to perform zero and few shot approaches have been discovered. These approaches propose the inclusion of a small number of supervised examples in the input as a hint for the model. The model then, just by looking a small set of examples, is able to complete successfully the task at hand. \citet{gpt-3} report that they solve a wide range of NLP tasks just following the previous approach. However, this approach only looks appropriate when the model is large enough.

In this paper we exploit the domain knowledge already encoded within the existing pre-trained Language Models to enrich the WordNet \citep{miller1998wordnet} synsets and glosses with domain labels. We explore and evaluate different pre-trained Language Models and pattern objectives. For instance, consider the example shown in Table \ref{tab:query-phrase}. Given a WordNet definition such as the one of $<$hospital, infirmary$>$ and the knowledge encoded in a pre-trained Language Model, the task is to assess which is its most suitable domain label. Thus, we create an appropriate pattern in natural language adapted to the objective of the Language Model. In the example, we use a Language Model fine-tuned on a general task such as Natural Language Inference (NLI) \cite{bowman2015large}. The NLI objective is to train a model able to classify the relation between two sentences as entailment, contradiction or neutral. Having four domains such as {\it medicine}, {\it biology}, {\it business} and {\it culture}, our system performs four queries to the model, each one with one of the four domains. Each query takes as a first sentence the WordNet definition and as a second sentence {\it The domain of the sentence is about [domain-label].} As expected, the most suitable domain label in this example is {\it medicine} with a confidence of 0.77. As shown, an off-the-shelf Language Model which have been fine-tuned on a general NLI task is able to infer the most appropriate domain label for the WordNet definition without any further training. Also note that the approach can use any given set of domain labels.

Interestingly, without any training on the task at hand, the proposed zero-shot system obtains an F1 score of 92.4\% on the English dataset used in the evaluation.

\begin{table*}[!ht]
    \centering
    \begin{tabularx}{0.77\linewidth}{lrclr}
        \toprule
        Definition: & \multicolumn{4}{c}{hospital: a health facility where patients receive treatment.} \\
        \midrule
        Pattern: & The domain of the sentence is about & \textbf{medicine} &   & \textbf{0.77} \\
                     &                                     & biology  &   & 0.08 \\
                     &                                     & business &   & 0.04 \\
                     &                                     & culture  &   & 0.02 \\
        \bottomrule
    \end{tabularx}
    \caption{An example of domain labelling.}
    \label{tab:query-phrase}
\end{table*}

All the implementation code along with the experiments is freely available on a GitHub repository \footnote{\url{https://github.com/osainz59/Ask2Transformers}}.

After this short introduction, the next section presents previous work on domain labelling of WordNet. Section \ref{methodology} presents our approach, Section \ref{experimental-setting} the experimental setup and Section \ref{evaluation-and-results} the results from our experiments. Finally, Section \ref{conclusions} revises the main conclusions and the future work.

\section{Related Work}
\label{related-work}

Building large and rich lexical knowledge bases is a very costly effort which involves large research groups for long periods of development. Starting from version 3.0, Princeton WordNet has associated topic information with a subset of its synsets. This topic labeling is achieved through pointers from a source synset to a target synset representing the topic. WordNet uses 440 topics and the most frequent one is $<$law, jurisprudence$>$. 

In order to reduce the manual effort required, a few semi-automatic and fully automatic methods have been applied for associating domain labels to synsets. For instance, WordNet Domains\footnote{\url{http://wndomains.fbk.eu/}} (WND) is a lexical resource where synsets have been semi-automatically annotated with one or more domain labels from a set of 165 hierarchically organized domains \cite{magnini2000g, bentivogli2004revising}. The uses of WND include the possibility to reduce the polysemy degree of the words, grouping those senses that belong to the same domain \cite{magnini2002role}. But the semi-automatic method used to develop this resource was far from being perfect. For instance, the noun synset $<$diver, frogman, underwater diver$>$ defined as {\it some-one who works underwater} has domain {\it history} because it inherits from its hypernym $<$explorer, adventurer$>$ also labelled with {\it history}. Moreover, many synsets have been labelled as {\it factotum} meaning that the synset cannot be labelled with a particular domain. WND also provides mappings to WordNet Topics and also to Wikipedia categories.

eXtended WordNet Domains\footnote{\url{https://adimen.si.ehu.es/web/XWND}} (XWND) \cite{gonzalez2012proposal,gonzalez2012graph} applied a graph-based method to propagate the WND labels through the WordNet structure.

Domain information is also available in other lexical resources. For instance, IATE\footnote{\url{http://iate.europa.eu/}}, a European Union inter-institutional terminology database. The domain labels of IATE are based on the Eurovoc thesaurus\footnote{\url{https://op.europa.eu/en/web/eu-vocabularies/th-dataset/-/resource/dataset/eurovoc}} and were introduced manually.



More recently, BabelDomains\footnote{\url{http://lcl.uniroma1.it/babeldomains/}} \cite{camacho2017} propose an automatic method that propagates the knowledge categories from the Wikipedia to WordNet by exploiting both distributional and graph-based clues. As  domains  of  knowledge,  BabelDomains  opted for  domains from the Wikipedia featured articles page\footnote{\url{https://en.wikipedia.org/wiki/Wikipedia:Featured_articles}}. This  page  contains  a  set  of  thirty-two domains of knowledge. When labelling WordNet synsets with these domains, BabelDomains reports a precision of 81.7, a recall of 68.7 and an F1 score of 74.6. Unfortunately, as these numbers suggest not all WordNet synsets have been labelled with a domain. For instance, the synset $<$hospital, infirmary$>$ with a gloss definition {\it a health facility where patients receive treatment} has no Babeldomain assigned.

It is worth to note that all these methods depart from a particular set of domain labels (or categories) manually assigned to a set of WordNet synsets (or Wikipedia pages). Then, these labels are propagated through the WordNet structure following automatic or semi-automatic methods. In contrast, our zero-shot method does not require an initial manual annotation. Furthermore, it is not designed for a particular set of domain labels. That is, it can be applied to label from scratch any dictionary or lexical knowledge base (or wordnet) with distinct sets of domain labels.


\section{Using pre-trained LMs for domain labelling}
\label{methodology}

Recent studies such as the one of GPT-3 \cite{gpt-3} shows that when increasing the size of the model, the capacity to solve different tasks with just a few positive examples also increases (few-shot learning). However, very large Language Models also have important hardware requirements (i.e. large RAM GPUs). Thus, we decided to keep the size of the models used manageable with small hardware requirements.

The task where we focused on is the domain labelling of WordNet glosses. This task consist in the following. Given a WordNet gloss $g$ to predict the corresponding domain $d$ of the WordNet concept defined. In this paper, the domains are taken from BabelDomains \cite{camacho2017}. Supervised domain labelling can be solved as any other multiclass problem, where the output of the model is a class probability distribution. In our zero-shot experiments we did not modify any of the pre-trained models. We just reformulate the domain labelling task to match with the LMs training objective.


\subsection{Masked Language Modeling}
\label{masked-language-model}

The Masked Language Modeling (MLM) is a pre-training objective followed by models such as BERT \cite{bert} and RoBERTa \cite{ROBERTA}. This objective works as follows. Given a sequence of tokens $s = [t_1, t_2, ..., t_n]$, the sequence is first perturbed by replacing some of the tokens $t$ with an special token [MASK]. Then, the model is trained to recover the original sequence $s$ given the modified sequence $\hat s$. This denoising objective can be seen as an evolution for the contextual embeddings of the previous CBOW from word2vec \cite{cbow}.

For domain labelling, we have replaced the input for the model following the next pattern:

\begin{figure}[!ht]
    \centering
    \begin{displayquote}
        $s$: Context: [context] Topic: [MASK]
    \end{displayquote}
    \label{fig:mlm_example}
\end{figure}

\noindent where we introduce the input sentence replacing the [context] tag. Then, we let the model predict the most probable token for the [MASK] tag. For instance, given the biological definition of {\it cell}, the model returns the following topics: {\it Biology}, {\it evolution}, {\it life}, etc.

This approach has been used to explore the knowledge of the model without any predefined set of domain labels in Section \ref{sseq:qualitative_analysis}.


\subsection{Next Sentence Prediction}

Along with the MLM the Next Sentence Prediction (NSP) is the training objective used by the BERT models. Given a pair of sentences $s_1$ and $s_2$, this objective predicts whether $s_1$ is followed by $s_2$ or not.

To adapt the BERT objective to the domain labelling task, we propose the next strategy inspired in the work from \newcite{yin-etal-2019-benchmarking}. We use the following input pattern:

\begin{figure}[!ht]
    \centering
    \begin{displayquote}
        $s_1$: [context] \\
        $s_2$: Domain or topic about [domain-label]
    \end{displayquote}
    \label{fig:nsp_example}
\end{figure}

\noindent where $s_1$ encodes a WordNet gloss as a context and $s_2$ is formed by a \textit{template} and a domain-label. In order to make the classification, we run as many times as domain labels and then apply a softmax over the positive class outputs. We hypothesize that, no matter if any of the $s_2$ can really follow the given $s_1$, the most probable one should be the $s_2$ formed by the correct label. For instance, recall the {\it hospital} example shown in Table \ref{tab:query-phrase}.

\subsection{Natural Language Inference}

In this case, we use a pre-trained LM that has been fine-tuned for a general inference task which is the Natural Language Inference \cite{MNLI}. Given two sentences in the form of a premise $s_1$ and an hypothesis $s_2$, the NLI task consists on redicting whether the $s_1$ \textit{entails} or \textit{contradicts} $s_2$ or if the relation between both is \textit{neutral}. 




We also used the input pattern shown in the previous NSP approach to adapt the NLI models to the domain labelling task. In this case, we just use the predictions of the \textit{entailment} class. The predictions of the {\textit contradiction} and \textit{neutral} are not used. As in the previous case, no matter if any of the $s_2$ hypothesis entails the premise $s_1$ or not, the most probable entailment should be the correct domain label. For example, consider again the example presented in Table \ref{tab:query-phrase}.


\section{Experimental setting}
\label{experimental-setting}

This section describes our experimental setup. We introduce the pre-trained Language Models and the dataset used. For the case of the Language Models, we have tested BERT \cite{bert}, RoBERTa \cite{ROBERTA} and BART \cite{BART}.  For the dataset, we have used the one released by \citet{camacho2016} based on WordNet.

\subsection{Pretrained models}

All the Language Models have been obtained from the Huggingface Transformers library \cite{transformers}.

\paragraph{MLM} For the objective we have used \textit{roberta-large} and \textit{roberta-base} checkpoints. These models have obtained state-of-the-art results on many NLP tasks and benchmarks. 

\paragraph{NSP} For this objective we use the BERT models as they are the only ones trained on that objective. For the sake of comparing the performance of more than one model of each objective we have selected the \textit{bert-large-uncased} and \textit{bert-base-uncased} checkpoints. They only differ on the size of the Language Model.

\paragraph{NLI} For this objective we used a checkpoint based on RoBERTa \textit{roberta-large-mnli} which have been fine-tuned with MultiNLI \cite{williams2018broad}. We also include \textit{bart-large-mnli} for testing a generative model.

\subsection{Dataset}

\begin{figure}
    \centering
    \includegraphics[scale=0.46]{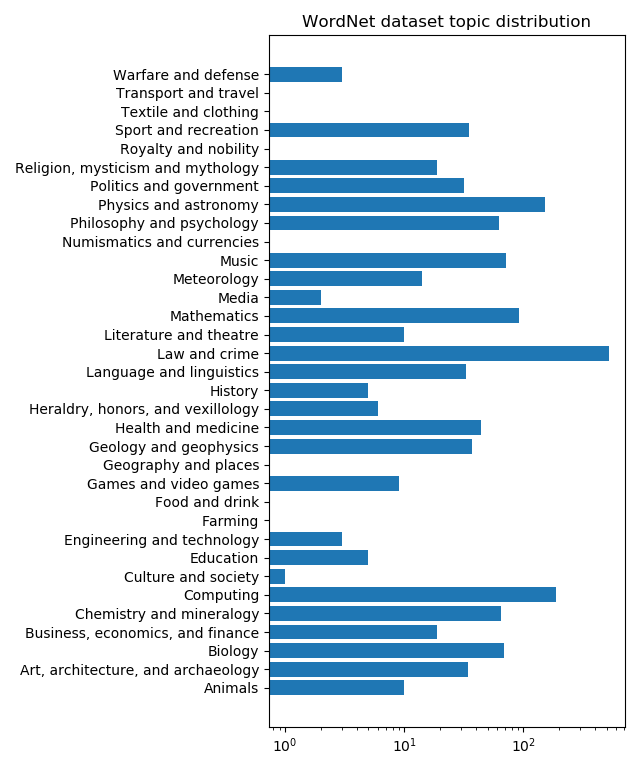}
    \caption{Distribution of domains in the WordNet dataset.}
    \label{fig:wordnet_topic_distribution}
\end{figure}

We evaluate our approaches on a dataset derived from WordNet which have been annotated with Babeldomain labels \cite{camacho2016}. This dataset consist of \textbf{1540} synsets manually annotated with their corresponding Babeldomain label. The distribution of domain labels in the dataset is shown in Figure \ref{fig:wordnet_topic_distribution}. Note that the dataset is quite unbalanced. In fact, some important domains such as {\it Transport and travel} or {\it Food and drink} have no single labelled example. As our system is unsupervised, we use the whole dataset for testing.

\section{Evaluation and Results}
\label{evaluation-and-results}

This section presents a quantitative and qualitative evaluation. One the one hand, the quantitative evaluation has been done incrementally in order to obtain the best-performing system. First, we have evaluated the different alternative models using the same objective pattern. Then, once the best approach was selected we have explored alternative patterns using the best model. When the  best performing pattern was discovered we have focus on finding a better label representation. Finally, we have compared our best system against the previous state-of-the-art methods.

On the other hand, as one of our system is based on a generative approach (MLM) the applied restrictions may not show the real performance of the method. So, we decided to at least do an small qualitative review of the approach.

\subsection{Approach comparison}

\begin{table}[!t]
    \centering
    \scalebox{.9}{
        \begin{tabular}{l|ccc}
            \toprule
             Method & Top-1 & Top-3 & Top-5 \\
            \midrule
            MNLI (roberta-large-mnli) & \textbf{78.44} & \textbf{87.46} & \textbf{89.74} \\ 
            MNLI (bart-large-mnli) & 61.81 & 79.85 & 87.59 \\
            NSP (bert-large-uncased) & 2.07 & 8.57 & 16.49 \\
            NSP (bert-base-uncased) & 2.85 & 10.32 & 16.88 \\
            \bottomrule
        \end{tabular}
    }
    
    \caption{Top-K accuracy of different approaches.}
    \label{tab:approach}
\end{table}

\begin{figure}[!t]
    \centering
    \includegraphics[scale=0.52]{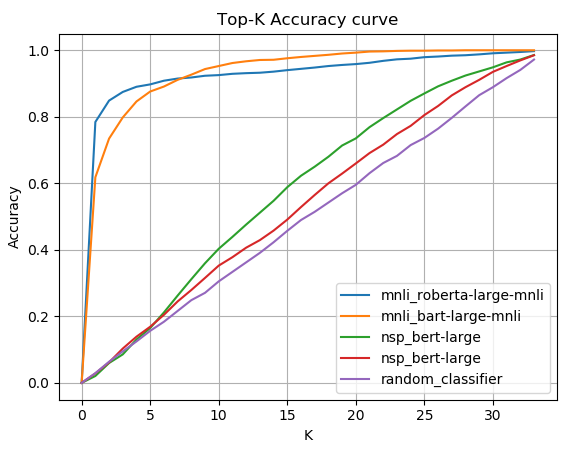}
    \caption{Top-K accuracy curve of the different approaches and a random classifier baseline.}
    \label{fig:topk_curve}
\end{figure}

Table \ref{tab:approach} shows the Top-1, Top-3 and Top-5 accuracy of each system when using the same objective pattern. To understand better the behaviour of the systems we also present in the Figure \ref{fig:topk_curve} the Top-K accuracy curve comparing all the approaches and a random baseline. As expected the systems that follow the same approaches perform similarly and share a similar curve. The best performing system is the MNLI based \textit{roberta-large-mnli}, followed by the \textit{bart-large-mnli} checkpoint. We observe a large difference between the different models. For instance, the models pre-trained on the NLI task perform much better than those pre-trained on the general NSP task.
The NSP approaches perform slightly better than the random classifier which can be a signal of a non appropriated objective model to use. 


\subsection{Input representation}

Once selected the pre-trained Language Model, we evaluate different input patterns for the \textit{roberta-large-mnli} checkpoint. As mentioned before, the MNLI approaches follow the same structure as NSP, where $s_1$ is the gloss of the synset and $s_2$ the sequence formed by a textual template plus the label.

\begin{table*}[!t]
    \centering
    \begin{tabularx}{0.77\linewidth}{lccc}
        \toprule
        Input pattern & Top-1 & Top-3 & Top-5 \\
        \midrule
        Topic: [label] & 59.61 & 69.48 & 74.02  \\
        Domain: [label] & 58.50 & 67.40 & 72.27 \\
        Theme: [label] & 59.67 & 73.96 & 81.36 \\
        Subject: [label] & 60.58 & 69.74 & 74.35 \\
        Is about [label] & 73.37 & 87.72 & 91.94 \\
        Topic or domain about [label] & 78.44 & 87.46 & 89.74 \\
        The topic of the sentence is about [label] & 80.71 & 92.92 & 95.77 \\
        The domain of the sentence is about [label] & \textbf{81.62} & \textbf{93.96} & \textbf{96.42} \\
        The topic or domain of the sentence is about [label] & 76.62 & 88.63 & 91.23 \\
        \bottomrule
    \end{tabularx}
    \caption{Some of the explored \textit{input patterns} for the MNLI approach and their Top-1, Top-3 and Top-5 accuracy.}
    \label{tab:query-phrase-exploration}
\end{table*}

Table \ref{tab:query-phrase-exploration} shows the results obtained by testing different textual patterns. Very short patterns obtain low results. The best performing textual template is obtained with \textit{The domain of the sentence is about [label]}.

\subsection{Label descriptors / Mapping}

As important as the input patterns is the set of domain labels used. Actually, BabelDomains uses labels that refers to one or several specific domains. For instance, {\it Art, architecture and archaeology}. Although these coarse-grained labels can be useful when clustering close-related domains, we also implemented a two-step labelling procedure taking into account those specific domains. First, we run the system over a set of specific domains or descriptors. Second, we apply a function that maps the descriptors to the original BabelDomains. 

\paragraph{Descriptors} The descriptors defined in this work are quite simple. Given a composed domain label such us {\it Art, architecture and archaeology}, we define the set of descriptors as each of the components of the label. For instance, in this case {\it Art}, {\it Architecture} and {\it Archaeology}. In the case of labels that consist on a single domain, the descriptors are just the labels. For example, in the case of {\it Music} the descriptor is also {\it Music}.

\paragraph{Mapping function} The mapping function that we use in this work consists on taking the maximum result of the descriptors as the result of the original domain label, i.e. $l_i = \max(d_{i1}, d_{i2}, ..., d_{in})$.

\subsection{Training a specialized student}

The inference time increases linearly with the number of labels. That is, for each example we need to test all the different domain labels. To speed-up the labelling process we annotate automatically the rest of WordNet glosses (around 79.000 glosses) using our best zero-shot approach. Then, we use that automatically annotated dataset to train a much smaller Language Model for the task. For instance, to label new definitions or new lexicons. We have fine-tuned two different models, the first one based with DistilBert \cite{sanh2019distilbert} which is 5 times smaller than the \textit{roberta-large-mnli} and a XLM-RoBERTa \cite{conneau2019unsupervised} \textit{base} which is 2 times smaller and is trained in a multilingual fashion. We called them A2T\textsubscript{FT-small} and A2T\textsubscript{FT-xlingual} respectively. The first one achieve a \textbf{x425} faster inference (5 times smaller and 85 times less inferences) while the second one a speed boost of \textbf{x170}.

\subsection{Results}

\begin{figure}[!ht]
    \centering
    \includegraphics[scale=.5]{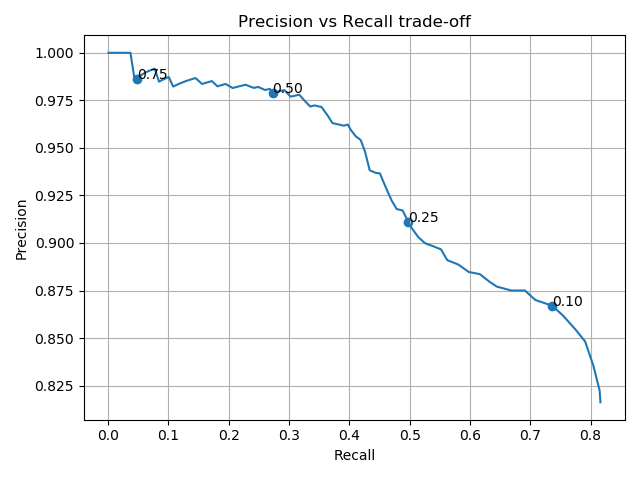}
    \caption{Precision/Recall trade-off of A2T system. Annotations indicates the probability thresholds.}
    \label{fig:precision-recall_tradeoff}
\end{figure}

In order to know how good is our final approach we compare our new systems with the previous ones. The results are reported on the Table \ref{tab:sota} in terms of Precision, Recall and F1 for comparison purposes. We also include the results from two previous state-of-the-art systems. As we can see, the new systems based on pre-trained Language Models obtain much better performance (from a previous best result with an F1 of 74.6 to the new one of 82.10). We also obtain an small improvement when
establishing a threshold to decide whether a prediction is taken into consideration or not. Our system performs slightly better with a confidence score greater than 5\%  (A2T\textsubscript{($> 0.05$)}). Figure \ref{fig:precision-recall_tradeoff} reports the Precision/Recall trade-off of the A2T system. As mentioned before labels composed of multiple domains can make the prediction harder for the zero-shot system. As a result, a simple system using the label descriptors boosts the performance of the system reaching a final \textbf{92.14} F1 score (A2T\textsubscript{+ descriptors}). Finally, we also include the results of both the  fine-tuned student versions which still obtain very competitive results while drastically reducing the inference time of the original models.

\begin{table}[!ht]
    \centering
    \begin{tabular}{l|ccc}
    \toprule
        Method & Precision & Recall & F1 \\
    \midrule
        Distributional & 84.0 & 59.8 & 69.9 \\
        BabelDomains & 81.7 & 68.7 & 74.6 \\
    \midrule
        A2T & 81.62 & 81.62 & 81.62 \\
        A2T\textsubscript{($> 0.05$)} & 83.20 & 81.03 & 82.10 \\
        A2T\textsubscript{+ descriptors} & \textbf{92.14} & \textbf{92.14} & \textbf{92.14} \\
    \midrule
        A2T\textsubscript{FT-small} & 91.42 & 91.42 & 91.42 \\
        A2T\textsubscript{FT-xlingual} & 90.58 & 90.58 & 90.58 \\
    \bottomrule
    \end{tabular}
    \caption{Micro-averaged precision, recall and F1 for each of the systems. Distributional \cite{camacho2016} and BabelDomains \cite{camacho2017} measures are the ones reported by them.}
    \label{tab:sota}
\end{table}

\begin{table*}[!ht]
    \centering
    \begin{tabularx}{0.9\linewidth}{l|cccc}
        \toprule
        Synset & cell & phase space & rounding error & wipeout\\
        \midrule
        Label & Biology & Physics and astronomy & Mathematics & Sports and Recreation \\
        \midrule
        Top         & \textbf{Biology}   & EOS           & rounding  & sports \\
        predictions & EOS       & physics       & EOS       & EOS \\
                    & biology   & \textbf{Physics}       & math      & sport \\
                    & evolution & geometry      & taxes     & accident \\
                    & life      & relativity    & \textbf{Math}      & \textbf{Sports}\\
        \bottomrule
    \end{tabularx}
    \caption{Top predictions of the MLM approach using the \textit{roberta-large} checkpoint.}
    \label{tab:mlm_examples}
\end{table*}


\subsection{Error analysis}

\begin{figure}
    \centering
    \includegraphics[scale=.35]{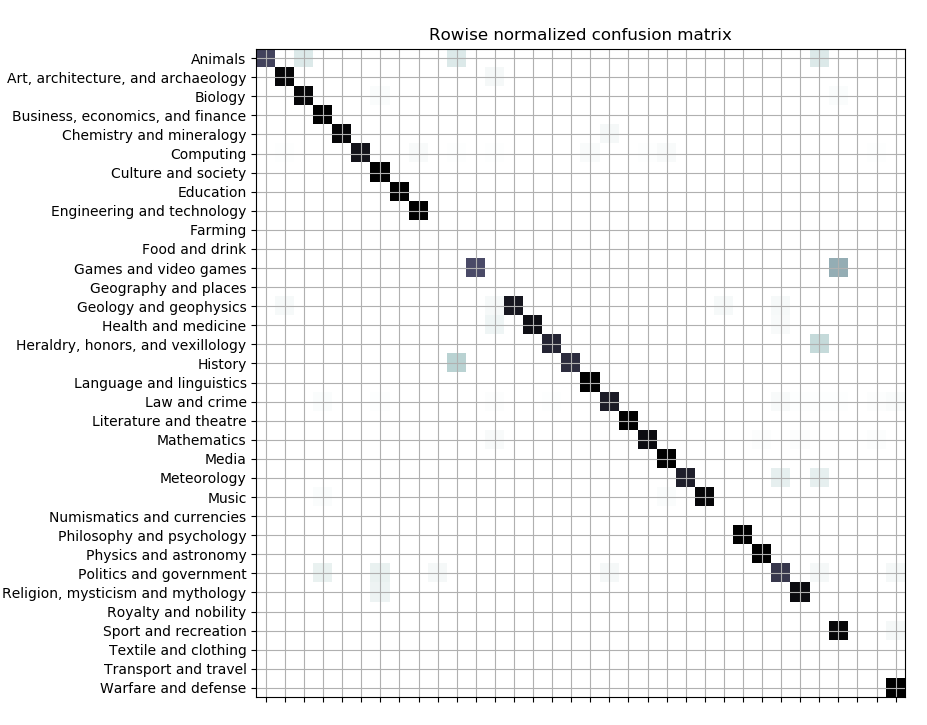}
    \caption{Rowise normalized confusion matrix of the A2T\textsubscript{+ descriptors} system.}
    \label{fig:confusion_matrix}
\end{figure}


Figure \ref{fig:confusion_matrix} presents the confusion matrix of our best system. The matrix is row wise normalized due to the imbalance of the dataset label distribution. Looking at the figure there are 4 classes that are misleading. The "Animals" domain is confused with the related domains "Biology" and "Food and drink". For instance, this is the case of the synset $<$diet$>$ with the definition {\it the usual food and drink consumed by an organism (person or animal)} which is labelled by our system as "Food and drink". The "Games and video games" domain is confused with the related domain "Sport and recreation". For example the sense referring to \textit{game: a single play of a sport or other contest; "the game lasted two hours"} which is labelled by our system as "Sport and recreation". The third one, "Heraldry, honors and vexillology" is also confused with a very close domain "Royalty and nobility". Obviously, close-related domains can be very difficult to distinguish even for humans. For example, the sense $<$audio cd, audio compact disc$>$ annotated in the gold standard as "Music" is labelled by our system as "Media".
Finally, sometimes the "History" domain is confused with "Food and drink". A curious example of this case is the sense referring to the history event $<$Boston tea party$>$ that is labelled as "Food and drink". 

\subsection{Qualitative analysis} \label{sseq:qualitative_analysis}

Table \ref{tab:mlm_examples} shows some of the top predictions obtained by a Masked Language Model (MLM) and the real label for 4 different synsets. In this case, the system is guessing its best predicted domain. That is, the system is not restricted to a select the best label from a pre-defined set of domain labels. Now, the system is free to return the word that best fit the masked term.


We can see in the table that the predictions of the model are  close to the correct label although not always equal. Sometimes because of a different case. They can also be seen as fine-grained domains or domain keywords of the real domain. 


\section{Conclusions and Future Work}
\label{conclusions}

In this paper we have explored some approaches for domain labelling of WordNet glosses by exploiting pre-trained LM in a zero-shot manner. We have presented a simple approach that achieves a new state-of the art on the Babeldomain dataset.

Even if we have focused on domain labelling of WordNet glosses, our method seems to be robust enough to be adapted to work on tasks such as Sentiment Analysis or other type of text classification. In particular, we think that the approach can be very useful when no annotated data is available.

For the future, we have considered three main objectives. First, we plan to apply this approach to other sources of domain information such as WordNet topics and WordNet Domains. We will also explore how to deal with definitions with generic domains (with no BabelDomains labels or with WordNet Domains factotum label). Second, we also aim to explore the cross-lingual capabilities of pre-trained Language Models for domain labelling of non-English wordnets and other lexical resources. Finally, we also plan to explore the utility of these findings in the Word Sense Disambiguation task.

\section*{Acknowledgments}

This work has been funded by the Spanish Ministry of Science, Innovation and Universities under the project DeepReading  (RTI2018-096846-B-C21)  (MCIU/AEI/FEDER,UE) and by the BBVA Big Data 2018 “BigKnowledge for Text Mining (BigKnowledge)” project. We  also  acknowledge  the  support  of  the  Nvidia Corporation with the donation of a GTX Titan X GPU used for this research.


\bibliography{bibliography}

\begin{thebibliography}{22}
\expandafter\ifx\csname natexlab\endcsname\relax\def\natexlab#1{#1}\fi

\bibitem[{Bentivogli et~al.(2004)Bentivogli, Forner, Magnini, and
  Pianta}]{bentivogli2004revising}
Luisa Bentivogli, Pamela Forner, Bernardo Magnini, and Emanuele Pianta. 2004.
\newblock Revising the wordnet domains hierarchy: semantics, coverage and
  balancing.
\newblock In \emph{Proceedings of the workshop on multilingual linguistic
  resources}, pages 94--101.

\bibitem[{Bowman et~al.(2015)Bowman, Angeli, Potts, and
  Manning}]{bowman2015large}
Samuel Bowman, Gabor Angeli, Christopher Potts, and Christopher~D Manning.
  2015.
\newblock A large annotated corpus for learning natural language inference.
\newblock In \emph{Proceedings of the 2015 Conference on Empirical Methods in
  Natural Language Processing}, pages 632--642.

\bibitem[{Brown et~al.(2020)Brown, Mann, Ryder, Subbiah, Kaplan, Dhariwal,
  Neelakantan, Shyam, Sastry, Askell et~al.}]{gpt-3}
Tom~B Brown, Benjamin Mann, Nick Ryder, Melanie Subbiah, Jared Kaplan, Prafulla
  Dhariwal, Arvind Neelakantan, Pranav Shyam, Girish Sastry, Amanda Askell,
  et~al. 2020.
\newblock \href {https://arxiv.org/pdf/2005.14165.pdf} {Language models are
  few-shot learners}.
\newblock \emph{arXiv preprint arXiv:2005.14165}.

\bibitem[{Camacho-Collados and Navigli(2017)}]{camacho2017}
Jose Camacho-Collados and Roberto Navigli. 2017.
\newblock \href {https://www.aclweb.org/anthology/E17-2036} {{B}abel{D}omains:
  Large-scale domain labeling of lexical resources}.
\newblock In \emph{Proceedings of the 15th Conference of the {E}uropean Chapter
  of the Association for Computational Linguistics: Volume 2, Short Papers},
  pages 223--228, Valencia, Spain. Association for Computational Linguistics.

\bibitem[{Camacho-Collados et~al.(2016)Camacho-Collados, Pilehvar, and
  Navigli}]{camacho2016}
José Camacho-Collados, Mohammad~Taher Pilehvar, and Roberto Navigli. 2016.
\newblock \href {https://doi.org/https://doi.org/10.1016/j.artint.2016.07.005}
  {Nasari: Integrating explicit knowledge and corpus statistics for a
  multilingual representation of concepts and entities}.
\newblock \emph{Artificial Intelligence}, 240:36 -- 64.

\bibitem[{Conneau et~al.(2020)Conneau, Khandelwal, Goyal, Chaudhary, Wenzek,
  Guzm{\'a}n, Grave, Ott, Zettlemoyer, and Stoyanov}]{conneau2019unsupervised}
Alexis Conneau, Kartikay Khandelwal, Naman Goyal, Vishrav Chaudhary, Guillaume
  Wenzek, Francisco Guzm{\'a}n, Edouard Grave, Myle Ott, Luke Zettlemoyer, and
  Veselin Stoyanov. 2020.
\newblock \href {https://doi.org/10.18653/v1/2020.acl-main.747} {Unsupervised
  cross-lingual representation learning at scale}.
\newblock In \emph{Proceedings of the 58th Annual Meeting of the Association
  for Computational Linguistics}, pages 8440--8451, Online. Association for
  Computational Linguistics.

\bibitem[{Devlin et~al.(2019)Devlin, Chang, Lee, and Toutanova}]{bert}
Jacob Devlin, Ming-Wei Chang, Kenton Lee, and Kristina Toutanova. 2019.
\newblock \href {https://doi.org/10.18653/v1/N19-1423} {{BERT}: Pre-training of
  deep bidirectional transformers for language understanding}.
\newblock In \emph{Proceedings of the 2019 Conference of the North {A}merican
  Chapter of the Association for Computational Linguistics: Human Language
  Technologies, Volume 1 (Long and Short Papers)}, pages 4171--4186,
  Minneapolis, Minnesota. Association for Computational Linguistics.

\bibitem[{Gonz{\'a}lez et~al.(2012)Gonz{\'a}lez, Rigau, and
  Castillo}]{gonzalez2012graph}
Aitor Gonz{\'a}lez, German Rigau, and Mauro Castillo. 2012.
\newblock A graph-based method to improve wordnet domains.
\newblock In \emph{International Conference on Intelligent Text Processing and
  Computational Linguistics}, pages 17--28. Springer.

\bibitem[{Gonzalez-Agirre et~al.(2012)Gonzalez-Agirre, Castillo, and
  Rigau}]{gonzalez2012proposal}
Aitor Gonzalez-Agirre, Mauro Castillo, and German Rigau. 2012.
\newblock A proposal for improving wordnet domains.
\newblock In \emph{LREC}, pages 3457--3462.

\bibitem[{Liu et~al.(2019)Liu, Ott, Goyal, Du, Joshi, Chen, Levy, Lewis,
  Zettlemoyer, and Stoyanov}]{ROBERTA}
Yinhan Liu, Myle Ott, Naman Goyal, Jingfei Du, Mandar Joshi, Danqi Chen, Omer
  Levy, Mike Lewis, Luke Zettlemoyer, and Veselin Stoyanov. 2019.
\newblock \href {https://arxiv.org/abs/1907.11692} {Roberta: A robustly
  optimized bert pretraining approach}.
\newblock \emph{arXiv preprint arXiv:1907.11692}.

\bibitem[{Magnini(2000)}]{magnini2000g}
B~Magnini. 2000.
\newblock G. cavagli a. integrating subject field codes into wordnet.
\newblock In \emph{Proceedings of LREC-2000, 2nd International Conference on
  Language Resources and Evaluation}, pages 1413--1418.

\bibitem[{Magnini et~al.(2002)Magnini, Strapparava, Pezzulo, and
  Gliozzo}]{magnini2002role}
Bernardo Magnini, Carlo Strapparava, Giovanni Pezzulo, and Alfio Gliozzo. 2002.
\newblock The role of domain information in word sense disambiguation.
\newblock \emph{Natural Language Engineering}, 8(4):359--373.

\bibitem[{Mikolov et~al.(2013)Mikolov, Chen, Corrado, and Dean}]{cbow}
Tomas Mikolov, Kai Chen, Greg Corrado, and Jeffrey Dean. 2013.
\newblock \href {https://arxiv.org/pdf/1301.3781.pdf} {Efficient estimation of
  word representations in vector space}.
\newblock \emph{arXiv preprint arXiv:1301.3781}.

\bibitem[{Miller(1998)}]{miller1998wordnet}
George~A Miller. 1998.
\newblock \emph{WordNet: An electronic lexical database}.
\newblock MIT press.

\bibitem[{Peters et~al.(2018)Peters, Neumann, Iyyer, Gardner, Clark, Lee, and
  Zettlemoyer}]{elmo}
Matthew Peters, Mark Neumann, Mohit Iyyer, Matt Gardner, Christopher Clark,
  Kenton Lee, and Luke Zettlemoyer. 2018.
\newblock \href {https://doi.org/10.18653/v1/N18-1202} {Deep contextualized
  word representations}.
\newblock In \emph{Proceedings of the 2018 Conference of the North {A}merican
  Chapter of the Association for Computational Linguistics: Human Language
  Technologies, Volume 1 (Long Papers)}, pages 2227--2237, New Orleans,
  Louisiana. Association for Computational Linguistics.

\bibitem[{Radford et~al.(2019)Radford, Wu, Child, Luan, Amodei, and
  Sutskever}]{gpt-2}
Alec Radford, Jeffrey Wu, Rewon Child, David Luan, Dario Amodei, and Ilya
  Sutskever. 2019.
\newblock Language models are unsupervised multitask learners.
\newblock \emph{OpenAI blog}, 1(8):9.

\bibitem[{Sanh et~al.(2019)Sanh, Debut, Chaumond, and
  Wolf}]{sanh2019distilbert}
Victor Sanh, Lysandre Debut, Julien Chaumond, and Thomas Wolf. 2019.
\newblock Distilbert, a distilled version of bert: smaller, faster, cheaper and
  lighter.
\newblock \emph{arXiv preprint arXiv:1910.01108}.

\bibitem[{Wang et~al.(2019)Wang, Zhao, Jia, Li, and Liu}]{BART}
Liang Wang, Wei Zhao, Ruoyu Jia, Sujian Li, and Jingming Liu. 2019.
\newblock \href {https://doi.org/10.18653/v1/D19-1412} {Denoising based
  sequence-to-sequence pre-training for text generation}.
\newblock In \emph{Proceedings of the 2019 Conference on Empirical Methods in
  Natural Language Processing and the 9th International Joint Conference on
  Natural Language Processing (EMNLP-IJCNLP)}, pages 4003--4015, Hong Kong,
  China. Association for Computational Linguistics.

\bibitem[{Williams et~al.(2018{\natexlab{a}})Williams, Nangia, and
  Bowman}]{MNLI}
Adina Williams, Nikita Nangia, and Samuel Bowman. 2018{\natexlab{a}}.
\newblock \href {http://aclweb.org/anthology/N18-1101} {A broad-coverage
  challenge corpus for sentence understanding through inference}.
\newblock In \emph{Proceedings of the 2018 Conference of the North American
  Chapter of the Association for Computational Linguistics: Human Language
  Technologies, Volume 1 (Long Papers)}, pages 1112--1122. Association for
  Computational Linguistics.

\bibitem[{Williams et~al.(2018{\natexlab{b}})Williams, Nangia, and
  Bowman}]{williams2018broad}
Adina Williams, Nikita Nangia, and Samuel Bowman. 2018{\natexlab{b}}.
\newblock A broad-coverage challenge corpus for sentence understanding through
  inference.
\newblock In \emph{Proceedings of the 2018 Conference of the North American
  Chapter of the Association for Computational Linguistics: Human Language
  Technologies, Volume 1 (Long Papers)}, pages 1112--1122.

\bibitem[{Wolf et~al.(2019)Wolf, Debut, Sanh, Chaumond, Delangue, Moi, Cistac,
  Rault, Louf, Funtowicz, and Brew}]{transformers}
Thomas Wolf, Lysandre Debut, Victor Sanh, Julien Chaumond, Clement Delangue,
  Anthony Moi, Pierric Cistac, Tim Rault, R'emi Louf, Morgan Funtowicz, and
  Jamie Brew. 2019.
\newblock \href {https://arxiv.org/pdf/1910.03771.pdf} {Huggingface's
  transformers: State-of-the-art natural language processing}.
\newblock \emph{ArXiv}, abs/1910.03771.

\bibitem[{Yin et~al.(2019)Yin, Hay, and Roth}]{yin-etal-2019-benchmarking}
Wenpeng Yin, Jamaal Hay, and Dan Roth. 2019.
\newblock \href {https://doi.org/10.18653/v1/D19-1404} {Benchmarking zero-shot
  text classification: Datasets, evaluation and entailment approach}.
\newblock In \emph{Proceedings of the 2019 Conference on Empirical Methods in
  Natural Language Processing and the 9th International Joint Conference on
  Natural Language Processing (EMNLP-IJCNLP)}, pages 3914--3923, Hong Kong,
  China. Association for Computational Linguistics.

\end{thebibliography}
\bibliographystyle{acl_natbib}

\end{document}